\documentclass[10pt,twocolumn]{article}
\usepackage[margin=0.8in]{geometry}
\usepackage{amsmath,amssymb,amsthm}
\usepackage{graphicx}
\usepackage{booktabs}
\usepackage{xcolor}
\usepackage[numbers,sort&compress]{natbib}
\usepackage[colorlinks=true,linkcolor=blue!50!black,citecolor=blue!50!black,urlcolor=blue!50!black]{hyperref}
\usepackage{microtype}

\expandafter\def\csname num@fullMs\endcsname{97}
\expandafter\def\csname num@eagerFullMs\endcsname{99}
\expandafter\def\csname num@mlpShare\endcsname{73}
\expandafter\def\csname num@eagerOnce20\endcsname{1.50}
\expandafter\def\csname num@eagerOnce30\endcsname{1.47}
\expandafter\def\csname num@eagerOnce40\endcsname{1.21}
\expandafter\def\csname num@eagerOnce60\endcsname{1.13}
\expandafter\def\csname num@chInterp\endcsname{3.1}
\expandafter\def\csname num@onceDev20\endcsname{2.44}
\expandafter\def\csname num@onceX20\endcsname{1.61}
\expandafter\def\csname num@perX20\endcsname{1.12}
\expandafter\def\csname num@onceOps20\endcsname{0.32}
\expandafter\def\csname num@perfRatio20\endcsname{1.36}
\expandafter\def\csname num@onceDev30\endcsname{2.00}
\expandafter\def\csname num@onceX30\endcsname{1.56}
\expandafter\def\csname num@perX30\endcsname{1.09}
\expandafter\def\csname num@onceOps30\endcsname{0.41}
\expandafter\def\csname num@perfRatio30\endcsname{1.34}
\expandafter\def\csname num@onceDev40\endcsname{1.57}
\expandafter\def\csname num@onceX40\endcsname{1.26}
\expandafter\def\csname num@perX40\endcsname{0.93}
\expandafter\def\csname num@onceOps40\endcsname{0.49}
\expandafter\def\csname num@perfRatio40\endcsname{1.42}
\expandafter\def\csname num@onceDev60\endcsname{0.78}
\expandafter\def\csname num@onceX60\endcsname{1.18}
\expandafter\def\csname num@perX60\endcsname{0.87}
\expandafter\def\csname num@onceOps60\endcsname{0.67}
\expandafter\def\csname num@perfRatio60\endcsname{1.74}
\expandafter\def\csname num@noneDev\endcsname{8.9}
\expandafter\def\csname num@noneX\endcsname{2.68}
\expandafter\def\csname num@chDev40\endcsname{5.13}
\expandafter\def\csname num@chX40\endcsname{1.84}
\expandafter\def\csname num@chDev50\endcsname{3.93}
\expandafter\def\csname num@chX50\endcsname{1.72}
\expandafter\def\csname num@chDev75\endcsname{1.83}
\expandafter\def\csname num@chX75\endcsname{1.29}
\expandafter\def\csname num@staticGapLo\endcsname{2.7}
\expandafter\def\csname num@staticGapHi\endcsname{2.9}
\expandafter\def\csname num@perfLo\endcsname{1.31}
\expandafter\def\csname num@perfHi\endcsname{1.74}
\expandafter\def\csname num@clFull\endcsname{10/10}
\expandafter\def\csname num@clThirty\endcsname{10/10}
\expandafter\def\csname num@clSixty\endcsname{9/10}
\expandafter\def\csname num@clNone\endcsname{0/10}
\expandafter\def\csname num@seedFlip\endcsname{16}
\expandafter\def\csname num@dRemRes\endcsname{47}
\expandafter\def\csname num@dRemPerf\endcsname{58}
\expandafter\def\csname num@dWristRes\endcsname{48}
\expandafter\def\csname num@dWristPerf\endcsname{41}
\expandafter\def\csname num@dAgentRes\endcsname{17}
\expandafter\def\csname num@dAgentPerf\endcsname{28}
\expandafter\def\csname num@onlyWrist\endcsname{3.0}
\expandafter\def\csname num@onlyAgent\endcsname{1.5}
\expandafter\def\csname num@onlyText\endcsname{2.9}
\expandafter\def\csname num@moveCopy\endcsname{14}
\expandafter\def\csname num@moveNorm\endcsname{36}
\expandafter\def\csname num@moveSel\endcsname{20}
\expandafter\def\csname num@rowsFrac\endcsname{32}
\expandafter\def\csname num@nObs\endcsname{30}
\newcommand{\fullMs}{\csname num@fullMs\endcsname}
\newcommand{\eagerFullMs}{\csname num@eagerFullMs\endcsname}
\newcommand{\mlpShare}{\csname num@mlpShare\endcsname}
\newcommand{\chInterp}{\csname num@chInterp\endcsname}
\newcommand{\noneDev}{\csname num@noneDev\endcsname}
\newcommand{\noneX}{\csname num@noneX\endcsname}
\newcommand{\staticGapLo}{\csname num@staticGapLo\endcsname}
\newcommand{\staticGapHi}{\csname num@staticGapHi\endcsname}
\newcommand{\perfLo}{\csname num@perfLo\endcsname}
\newcommand{\perfHi}{\csname num@perfHi\endcsname}
\newcommand{\clFull}{\csname num@clFull\endcsname}
\newcommand{\clThirty}{\csname num@clThirty\endcsname}
\newcommand{\clSixty}{\csname num@clSixty\endcsname}
\newcommand{\clNone}{\csname num@clNone\endcsname}
\newcommand{\seedFlip}{\csname num@seedFlip\endcsname}
\newcommand{\dRemRes}{\csname num@dRemRes\endcsname}
\newcommand{\dRemPerf}{\csname num@dRemPerf\endcsname}
\newcommand{\dWristRes}{\csname num@dWristRes\endcsname}
\newcommand{\dWristPerf}{\csname num@dWristPerf\endcsname}
\newcommand{\dAgentRes}{\csname num@dAgentRes\endcsname}
\newcommand{\dAgentPerf}{\csname num@dAgentPerf\endcsname}
\newcommand{\onlyWrist}{\csname num@onlyWrist\endcsname}
\newcommand{\onlyAgent}{\csname num@onlyAgent\endcsname}
\newcommand{\onlyText}{\csname num@onlyText\endcsname}
\newcommand{\moveCopy}{\csname num@moveCopy\endcsname}
\newcommand{\moveNorm}{\csname num@moveNorm\endcsname}
\newcommand{\moveSel}{\csname num@moveSel\endcsname}
\newcommand{\rowsFrac}{\csname num@rowsFrac\endcsname}
\newcommand{\nObs}{\csname num@nObs\endcsname}
\newcommand{\onceDev}[1]{\csname num@onceDev#1\endcsname}
\newcommand{\onceX}[1]{\csname num@onceX#1\endcsname}
\newcommand{\perX}[1]{\csname num@perX#1\endcsname}
\newcommand{\eagerOnce}[1]{\csname num@eagerOnce#1\endcsname}
\newcommand{\onceOps}[1]{\csname num@onceOps#1\endcsname}
\newcommand{\perfRatio}[1]{\csname num@perfRatio#1\endcsname}
\newcommand{\chDev}[1]{\csname num@chDev#1\endcsname}
\newcommand{\chX}[1]{\csname num@chX#1\endcsname}

\expandafter\def\csname num@locPerf\endcsname{58}
\newcommand{\locPerf}{\csname num@locPerf\endcsname}
\expandafter\def\csname num@locNov\endcsname{47}
\newcommand{\locNov}{\csname num@locNov\endcsname}
\expandafter\def\csname num@locInAbs\endcsname{53}
\newcommand{\locInAbs}{\csname num@locInAbs\endcsname}
\expandafter\def\csname num@devPerfB\endcsname{2.62}
\newcommand{\devPerfB}{\csname num@devPerfB\endcsname}
\expandafter\def\csname num@devNovB\endcsname{2.00}
\newcommand{\devNovB}{\csname num@devNovB\endcsname}
\expandafter\def\csname num@decDense\endcsname{194}
\newcommand{\decDense}{\csname num@decDense\endcsname}
\expandafter\def\csname num@decThirty\endcsname{160}
\newcommand{\decThirty}{\csname num@decThirty\endcsname}
\expandafter\def\csname num@decSixty\endcsname{181}
\newcommand{\decSixty}{\csname num@decSixty\endcsname}
\expandafter\def\csname num@decXThirty\endcsname{1.22}
\newcommand{\decXThirty}{\csname num@decXThirty\endcsname}
\expandafter\def\csname num@decXSixty\endcsname{1.08}
\newcommand{\decXSixty}{\csname num@decXSixty\endcsname}
\expandafter\def\csname num@decPrefDense\endcsname{94.5}
\newcommand{\decPrefDense}{\csname num@decPrefDense\endcsname}
\expandafter\def\csname num@decPrefThirty\endcsname{60.1}
\newcommand{\decPrefThirty}{\csname num@decPrefThirty\endcsname}
\expandafter\def\csname num@decImage\endcsname{34.8}
\newcommand{\decImage}{\csname num@decImage\endcsname}
\expandafter\def\csname num@decExpert\endcsname{62.0}
\newcommand{\decExpert}{\csname num@decExpert\endcsname}
\expandafter\def\csname num@decPrefShare\endcsname{49}
\newcommand{\decPrefShare}{\csname num@decPrefShare\endcsname}
\expandafter\def\csname num@peakTflops\endcsname{36.9}
\newcommand{\peakTflops}{\csname num@peakTflops\endcsname}
\expandafter\def\csname num@bwGbps\endcsname{342}
\newcommand{\bwGbps}{\csname num@bwGbps\endcsname}
\expandafter\def\csname num@ridge\endcsname{108}
\newcommand{\ridge}{\csname num@ridge\endcsname}
\expandafter\def\csname num@prefPeak\endcsname{82}
\newcommand{\prefPeak}{\csname num@prefPeak\endcsname}
\expandafter\def\csname num@expBw\endcsname{51}
\newcommand{\expBw}{\csname num@expBw\endcsname}
\expandafter\def\csname num@lowPeak\endcsname{42}
\newcommand{\lowPeak}{\csname num@lowPeak\endcsname}
\expandafter\def\csname num@rowsPeak\endcsname{67}
\newcommand{\rowsPeak}{\csname num@rowsPeak\endcsname}
\expandafter\def\csname num@prefBw\endcsname{18}
\newcommand{\prefBw}{\csname num@prefBw\endcsname}
\expandafter\def\csname num@prefIntensity\endcsname{484}
\newcommand{\prefIntensity}{\csname num@prefIntensity\endcsname}
\expandafter\def\csname num@expIntensity\endcsname{47}
\newcommand{\expIntensity}{\csname num@expIntensity\endcsname}
\expandafter\def\csname num@wamN\endcsname{8}
\newcommand{\wamN}{\csname num@wamN\endcsname}
\expandafter\def\csname num@wNov\endcsname{12.15}
\newcommand{\wNov}{\csname num@wNov\endcsname}
\expandafter\def\csname num@wPerf\endcsname{12.60}
\newcommand{\wPerf}{\csname num@wPerf\endcsname}
\expandafter\def\csname num@wPeek\endcsname{3.22}
\newcommand{\wPeek}{\csname num@wPeek\endcsname}
\expandafter\def\csname num@wCurPerf\endcsname{11.19}
\newcommand{\wCurPerf}{\csname num@wCurPerf\endcsname}
\expandafter\def\csname num@wCurNov\endcsname{13.07}
\newcommand{\wCurNov}{\csname num@wCurNov\endcsname}
\expandafter\def\csname num@wPrevPerf\endcsname{5.16}
\newcommand{\wPrevPerf}{\csname num@wPrevPerf\endcsname}
\expandafter\def\csname num@wPrevNov\endcsname{6.32}
\newcommand{\wPrevNov}{\csname num@wPrevNov\endcsname}
\expandafter\def\csname num@wDownPerf\endcsname{3.33}
\newcommand{\wDownPerf}{\csname num@wDownPerf\endcsname}
\expandafter\def\csname num@wDownNov\endcsname{3.91}
\newcommand{\wDownNov}{\csname num@wDownNov\endcsname}
\expandafter\def\csname num@wDownNovX\endcsname{0.33}
\newcommand{\wDownNovX}{\csname num@wDownNovX\endcsname}
\expandafter\def\csname num@wDownNovWins\endcsname{8}
\newcommand{\wDownNovWins}{\csname num@wDownNovWins\endcsname}
\expandafter\def\csname num@wDownVsPrevX\endcsname{0.65}
\newcommand{\wDownVsPrevX}{\csname num@wDownVsPrevX\endcsname}
\expandafter\def\csname num@wDownVsPrevWins\endcsname{8}
\newcommand{\wDownVsPrevWins}{\csname num@wDownVsPrevWins\endcsname}
\expandafter\def\csname num@wDownVsLaqWins\endcsname{8}
\newcommand{\wDownVsLaqWins}{\csname num@wDownVsLaqWins\endcsname}
\expandafter\def\csname num@wPrevGaze\endcsname{9.44}
\newcommand{\wPrevGaze}{\csname num@wPrevGaze\endcsname}
\expandafter\def\csname num@wDownGaze\endcsname{4.87}
\newcommand{\wDownGaze}{\csname num@wDownGaze\endcsname}
\expandafter\def\csname num@wPeekGaze\endcsname{4.63}
\newcommand{\wPeekGaze}{\csname num@wPeekGaze\endcsname}
\expandafter\def\csname num@wNovGainPct\endcsname{21}
\newcommand{\wNovGainPct}{\csname num@wNovGainPct\endcsname}
\expandafter\def\csname num@wNovGainWins\endcsname{8}
\newcommand{\wNovGainWins}{\csname num@wNovGainWins\endcsname}
\expandafter\def\csname num@wGazeDownX\endcsname{0.57}
\newcommand{\wGazeDownX}{\csname num@wGazeDownX\endcsname}
\expandafter\def\csname num@wGazeDownWins\endcsname{8}
\newcommand{\wGazeDownWins}{\csname num@wGazeDownWins\endcsname}

\title{NoveltyGate: Robots Should Compute\\What They Haven't Seen}
\author{Lemeng Qi\\Johns Hopkins University}
\date{}

\begin{document}
\maketitle

\begin{abstract}
When a robot policy runs its MLPs through low-rank drafts and can afford exact computation for only a fraction of its tokens, which tokens should it choose? The answer is input novelty: how much of a token's MLP input lies outside the subspace seen during calibration. In the $\pi_{0.5}$ vision-language-action (VLA) prefix, novelty outperforms a verifier that knows every token's exact MLP error. At a 30\% exact-token budget the verifier removes \locPerf{}\% of squared MLP error versus \locNov{}\% for novelty, yet its action deviation is \devPerfB{} seed units versus \devNovB{}; normalizing either score by token magnitude does not change the outcome. Novelty wins because it points where the action reads: its ranking follows the action expert's later attention (rank correlation 0.59) far more closely than exact error does (0.34). Novelty stays essential in world--action models (WAMs), where the action's attention is visible during denoising. In Cosmos Policy, multiplying the previous step's downstream action attention by novelty lowers deviation by \wNovGainPct{}\% on every test observation, reaching \wDownNovX{}$\times$ the deviation of novelty alone. Because novelty is read off the MLP input, NoveltyGate decides once at each MLP entrance with a fixed quota, so the whole $\pi_{0.5}$ prefix runs as one CUDA graph. Those MLPs are compute-bound, and a 30\% budget cuts a whole control decision from \decDense{} to \decThirty{}\,ms on a laptop GPU. Like the original policy, NoveltyGate completes all ten paired LIBERO episodes, while an unchecked low-rank prefix completes none.
\end{abstract}

\section{Introduction}
At each control call, $\pi_{0.5}$ encodes its camera views and instruction into a prefix cache that a smaller expert reads to predict an action chunk~\cite{pi0,pi05}. The prefix MLPs account for about \mlpShare{}\% of prefix GPU time on our device. Replacing them with rank-256 drafts saves arithmetic, but running every token through the drafts causes all ten closed-loop episodes in our evaluation to fail. Suppose we can afford to send three tokens in ten through the original MLP. Which three should we choose?

\begin{figure}[!ht]
\centering
\includegraphics[width=\columnwidth]{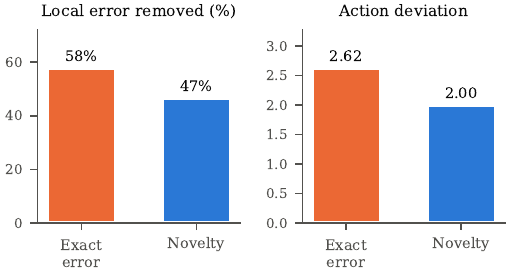}
\caption{\textbf{The ranking reverses at the action output.} Both rules keep 30\% of tokens exact, on the same 30 observations of held-out set B. Left: share of squared MLP draft error carried by the selected tokens, averaged over layers. Right: mean action deviation in seed units (lower is better).}
\label{fig:reversal}
\end{figure}

The obvious rule is to repair the largest mistakes. We give it every advantage: the exact error of each token's entire MLP output, at zero scoring cost. NoveltyGate asks a simpler question of each token's input: how much of it lies outside the subspace seen during calibration, the subspace the draft was fitted on. This novelty score is adapted from SSI/GSI~\cite{ssi,gsi}. The verifier removes more local error (\locPerf{}\% versus \locNov{}\%), yet novelty produces actions closer to those of the original policy (\devNovB{} versus \devPerfB{} seed units, Fig.~\ref{fig:reversal}). Novelty wins on two held-out sets and at every exact-token budget we test, and dividing either score by token magnitude does not change the outcome.

Novelty wins because it points where the action reads. Its ranking follows the attention that the action expert later pays to each prefix token with a rank correlation of 0.59, against 0.34 for exact error: the largest draft errors sit on tokens the action barely uses. Novelty recovers this preference from the input alone, before the action expert has run.

Novelty is also the signal a fast implementation needs. It is read off the 2048-wide MLP input, so NoveltyGate decides once per token and layer, before any wide product, and applies the decision to all three MLP products. Exact error is available only after the product it is meant to skip, and error sketches must be evaluated product by product. Deciding at the entrance raises the eager prefix speedup at a 20\% exact budget from \perX{20}$\times$ to \eagerOnce{20}$\times$, and a fixed quota gives the whole prefix static shapes, so it runs as one CUDA graph. Because the prefix MLPs are compute-bound, a 30\% budget cuts a whole control decision from \decDense{} to \decThirty{}\,ms, and NoveltyGate completes all ten paired closed-loop episodes.

Novelty stays essential in world--action models. There one transformer denoises the action frame together with video, so the action's attention is visible during dispatch, and we can ask whether novelty still matters once the reader is known. It does. On Cosmos Policy, NoveltyGate weights novelty by the previous denoising step's action attention from the blocks downstream of each MLP, the blocks that read its output. Novelty lowers the deviation of this attention alone by \wNovGainPct{}\% on every test observation, reaching \wDownNovX{}$\times$ the deviation of novelty alone, and the pair outperforms the token selection of concurrent LAQ~\cite{laq} even when LAQ's rule is given exact error (Sec.~\ref{sec:wam}).

We contribute: (i) the finding that input novelty allocates exact computation in a VLA prefix better than exact knowledge of every token's MLP error, because it tracks where the action reads; (ii) evidence that novelty remains essential in a WAM, where it improves on downstream-aligned action attention on every test observation; and (iii) NoveltyGate, which exploits novelty's availability at the MLP entrance to make one statically shaped decision per layer and turns the saved arithmetic into end-to-end time.

\section{Setting}
\paragraph{The prefix.} The released $\pi_{0.5}$ LIBERO checkpoint~\cite{libero} pads the prefix to 968 slots, of which about 660 are valid: 256 tokens per camera and the instruction with the discretized state. We use the strongest exact baseline we know: masked slots are removed, and the MLP of the last layer is skipped because no later layer reads its output. All speedups are measured against this prefix, which reproduces the native key-value cache bit for bit.

\begin{figure}[t]
\centering
\includegraphics[width=\columnwidth]{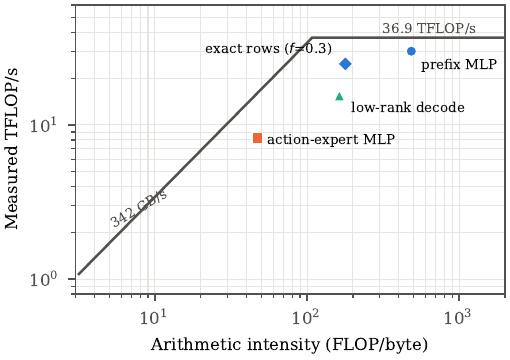}
\caption{Measured roofline of the RTX 5060 Laptop GPU at 115\,W (peak BF16 matrix throughput and copy bandwidth measured on the device). Points are batch-one products with the shapes of one $\pi_{0.5}$ control call.}
\label{fig:roofline}
\end{figure}

\paragraph{Where the time goes.} A batch-one control call on our RTX~5060 Laptop GPU takes \decDense{}\,ms: \decImage{}\,ms of image encoding, \decPrefDense{}\,ms of prefix (\decPrefShare{}\%) and \decExpert{}\,ms for the ten action-expert steps (Table~\ref{tab:decision}). The two parts of the language model sit on opposite sides of the roofline (Fig.~\ref{fig:roofline}). A prefix MLP product over 660 tokens has an arithmetic intensity of \prefIntensity{}\,FLOP/byte and runs at \prefPeak{}\% of the measured \peakTflops{}\,TFLOP/s peak while using \prefBw{}\% of the \bwGbps{}\,GB/s bandwidth. An action-expert MLP product over 50 tokens has intensity \expIntensity{} and uses \expBw{}\% of the bandwidth. Nsight Compute counters on the stock kernels agree: the three prefix MLP products keep the tensor pipes active 89--95\% of the time, while the three action-expert products reach 73--84\% of peak DRAM throughput. Reducing arithmetic therefore pays off in the prefix MLPs and not in the action expert, which is why we approximate the former. Smaller products run less efficiently: a rank-256 decode reaches \lowPeak{}\% of peak and the exact path over 30\% of the rows \rowsPeak{}\%, which is why saved arithmetic does not translate one-for-one into time.

\paragraph{Deviation.} For each observation we run the action expert with 4 noise seeds and compare each modified prefix to the original under identical noise. The deviation is the RMS difference of the six pose dimensions over the 50-step chunk, divided by the median RMS difference between two seeds of the original policy. A value below 1 means the modification moves the action less than resampling the noise does. We also report the fraction of steps whose gripper command keeps its sign.

\paragraph{Protocol.} Drafts are fitted on 31 observations from closed-loop LIBERO-10 rollouts of tasks 0, 1, 3, 5 and 7, and rules are compared on other initial states of the same tasks. Everything is then frozen and evaluated on tasks 2, 4, 6, 8 and 9, which never enter fitting or selection. Held-out set A (initial states 7 and 8) compared per-product rules; after it, we moved the decision to once per layer and evaluated on a fresh held-out set B (initial states 11 and 12) and in closed loop (initial states 13 and 14). Each held-out set has 30 observations, three per episode. Timing uses an RTX~5060 Laptop GPU on AC power (enforced power limit 115\,W, checked before every observation) with PyTorch~2.7 in BF16, on six held-out observations. For the main comparison, the whole prefix of every arm, including the original, is captured as a CUDA graph; each arm is replayed in random alternation with the original prefix, 72 pairs per arm, and we report the ratio of median times. Whole-decision timing (Table~\ref{tab:decision}) captures image encoding, prefix and the action-expert steps as separate CUDA graphs; to stay within the 8\,GB of device memory, each process holds the original prefix and a single dispatch prefix. The original prefix takes \fullMs{}\,ms in this mode and \eagerFullMs{}\,ms in eager execution, and the MLP accounts for about \mlpShare{}\% of it. Channel pruning is timed on weights whose intermediate channels are permuted by importance, so a pruned MLP is a contiguous slice of the original weights.

\section{Which tokens deserve exact computation?}
\subsection{A perfect local verifier loses on action fidelity}
Table~\ref{tab:rules} holds the draft and the exact fraction fixed and changes only the score. On set B the input residual and the union perform alike (within 5\% at every $f$, with intervals covering 1); on set A, with per-product decisions, the union is 4--7\% lower. Both are clearly better than the output sketch and the perfect verifier, so the output-error signal adds at most a few percent on top of the input residual and is never the better signal on its own. Relative to the input residual, the perfect verifier deviates by \perfRatio{20}, \perfRatio{30}, \perfRatio{40} and \perfRatio{60}$\times$ at $f = 0.2, 0.3, 0.4, 0.6$; on held-out set A, with one decision per product, the exact error of each product loses in the same way, and over both sets and all fractions the ratio lies between \perfLo{} and \perfHi{}, with every episode-bootstrap 95\% interval above 1 (Table~\ref{tab:robust}).

\begin{table}[t]
\centering
\small
\setlength{\tabcolsep}{3.5pt}
\begin{tabular}{lccccc}
\toprule
 & Input & Input $\cup$ & Output & Perfect & \\
$f$ & residual & output & sketch & verifier & Random \\
\midrule
$0.2$ & 2.44 & 2.44 & 3.35 & 3.25 & 6.72 \\
$0.3$ & 2.00 & 1.97 & 2.57 & 2.62 & 6.06 \\
$0.4$ & 1.57 & 1.51 & 1.99 & 2.11 & 5.04 \\
$0.6$ & 0.78 & 0.76 & 1.12 & 1.16 & 3.59 \\
\bottomrule
\end{tabular}
\caption{Which tokens to compute exactly. Mean deviation on held-out set B for one dispatch decision per token and layer, same draft, same exact fraction $f$. The perfect verifier knows the exact error of every token's MLP output and is charged nothing.}
\label{tab:rules}
\end{table}

\begin{table}[t]
\centering
\small
\setlength{\tabcolsep}{3pt}
\resizebox{\columnwidth}{!}{%
\begin{tabular}{llccc}
\toprule
Set & Decision & $f$ & Perfect verifier / residual & Union / residual \\
\midrule
A & per product & $0.2$ & 1.34 [1.18, 1.52] & 0.96 [0.88, 1.04] \\
A & per product & $0.4$ & 1.35 [1.22, 1.50] & 0.93 [0.88, 0.99] \\
A & per product & $0.6$ & 1.33 [1.12, 1.60] & 0.96 [0.84, 1.11] \\
B & per product & $0.3$ & 1.32 [1.15, 1.51] & -- \\
B & per product & $0.6$ & 1.31 [1.11, 1.49] & -- \\
B & per layer & $0.2$ & 1.36 [1.18, 1.55] & 1.00 [0.96, 1.04] \\
B & per layer & $0.3$ & 1.34 [1.14, 1.53] & 0.98 [0.95, 1.01] \\
B & per layer & $0.4$ & 1.42 [1.17, 1.67] & 0.96 [0.87, 1.08] \\
B & per layer & $0.6$ & 1.74 [1.30, 2.25] & 1.04 [0.98, 1.10] \\
\bottomrule
\end{tabular}}
\caption{Deviation ratios with episode-bootstrap 95\% intervals on the two held-out sets (30 observations each). A ratio above 1 means the score in the numerator picks worse tokens than the input residual. The perfect verifier ranks tokens by the exact error of the product (per-product decisions) or of the whole MLP (per-layer decisions).}
\label{tab:robust}
\end{table}

\paragraph{Is it normalization?} Novelty divides by the token's input norm, while the verifier ranks by absolute error, so a simpler explanation is that dividing by magnitude is what helps. Table~\ref{tab:fourcell} separates the two axes on the same observations. Dividing the output error by the output norm makes it worse, not better (\(2.72\) versus \(2.62\)), and both output scores lose to novelty with intervals excluding 1. The reversal also appears inside the input family: the absolute input residual removes more local error than novelty (\locInAbs{}\% versus \locNov{}\%) and produces larger action deviation. Across the four deterministic scores, removing more local error does not predict better actions.

\begin{table}[t]
\centering
\small
\setlength{\tabcolsep}{3pt}
\resizebox{\columnwidth}{!}{%
\begin{tabular}{lccc}
\toprule
 & Local error & Action & Deviation ratio \\
Score (30\% exact) & removed & deviation & vs.\ novelty \\
\midrule
Input residual, relative (novelty) & 46.9\% & 2.00 & 1.00 \\
Input residual, absolute & 53.4\% & 2.32 & 1.16 [1.05, 1.25] \\
Output error, absolute (perfect verifier) & 57.8\% & 2.62 & 1.34 [1.14, 1.52] \\
Output error, relative & 44.1\% & 2.72 & 1.35 [1.18, 1.58] \\
Random & 30.1\% & 6.16 & 3.05 [2.50, 3.59] \\
\bottomrule
\end{tabular}}
\caption{Input versus output, absolute versus relative. Same draft, same 30\% quota, one decision per layer, 30 observations of set B; local error is the share of squared MLP draft error carried by the selected tokens, computed on the same observations. Ratios are geometric means with episode-bootstrap 95\% intervals.}
\label{tab:fourcell}
\end{table}

\subsection{Where the action expert reads}
In a dense-input diagnostic on the first 12 observations of set B, the verifier selects \dAgentPerf{}\% of third-person tokens and \dWristPerf{}\% of wrist-camera tokens. Novelty selects \dAgentRes{}\% and \dWristRes{}\%, respectively. These percentages are selection rates \emph{within each camera}. Approximating just the wrist view on calibration observations moves the actions by \onlyWrist{} seed units, versus \onlyAgent{} for the third-person view. A fixed wrist-first modality priority nevertheless produces a deviation of 4.93 on set B at $f=0.3$, compared with 1.99 for novelty.

To test whether the missing information concerns the reader, we record attention from the action expert during dense reference denoising. For an MLP at layer $\ell$, we average prefix attention over seeds, heads and action positions and sum it over denoising steps and layers after $\ell$: these are the cache entries that its output can influence. The mean within-observation, within-layer Spearman correlation with this score is 0.59 for novelty and 0.34 for exact MLP error. On set B at $f=0.3$, ranking by reference attention alone yields 0.77 seed units. Multiplying it by the exact local error gives 0.58, compared with 2.62 for error alone. The weighted score beats novelty on all 30 observations in this diagnostic run. Its novelty baseline is 1.99 (2.00 in the main implementation in Table~\ref{tab:rules}).

Novelty thus recovers the action expert's token preference from the input alone. The reference attention exists only after the prefix has been computed, and its deployable stand-in, a position-wise attention map averaged on calibration data, gives 1.97 on the same set: no better than novelty, which adapts to each observation.

\subsection{Novelty when the reader is visible: world--action models}\label{sec:wam}
In $\pi_{0.5}$ the reader runs after the prefix, and novelty stands in for it. A world--action model lets us ask whether novelty still matters once the reader is known. In such a model the action is one frame of a video that a single transformer denoises jointly: action tokens attend to every token in every block at every denoising step, so the reader's attention is computed during the very forward pass we are dispatching. We test what this buys on Cosmos Policy 2B~\cite{cosmos} (28 DiT blocks, $9\times196=1764$ tokens, five denoising steps), with rank-256 output-optimal drafts per block fitted on three trajectories, 30\% of the tokens exact in every block and step, and \wamN{} observations from two held-out trajectories. Deviation is the mean absolute action difference to a dense run, divided by the difference between two dense seeds; all Cosmos numbers below use this unit.

\paragraph{Novelty adds to the best available reader.} The MLP of block $b$ writes into the residual stream that blocks $b{+}1,\dots,B$ read, so its readers are downstream. The previous denoising step supplies them: we record the action rows' attention in each block of the dispatched run and, at the next step, weight block $b$'s tokens by the attention summed over blocks after $b$; step 0 uses the mean attention of three calibration observations. Used alone, this attention gives \wDownGaze{} (Table~\ref{tab:wam}). Multiplying it by novelty gives \wDownNov{}, a \wNovGainPct{}\% reduction that holds on all \wNovGainWins{} observations, and \wDownNovX{}$\times$ the deviation of novelty alone. The rule uses only quantities available at inference. Novelty on its own matches an exact-error oracle at this budget (\wNov{} versus \wPerf{}; the oracle has the lower mean at 20\%, 40\% and 60\%), and the reader multiplies its value. Attention of block $b$ itself, which is available before its MLP but does not read its output, adds nothing (\wCurNov{} with novelty), and a dense reference run bounds what a reader can add (\wPeekGaze{} alone, \wPeek{} with exact error).

\paragraph{Against LAQ's token selection.} The closest token-selection rule is that of concurrent LAQ~\cite{laq}, which weights tokens by previous-step action attention summed over all layers and multiplies by an error estimate. We transplant it into the same router and give it exact error in place of its variance estimate. It reaches \wPrevPerf{}; NoveltyGate's downstream rule reaches \wDownNov{} with no exact error, better on all \wDownVsLaqWins{} observations. Two ingredients account for the gap, and each wins on every observation: novelty, which sharpens the aligned attention by \wNovGainPct{}\% (\wDownGaze{} to \wDownNov{}), and aligning the reader with the blocks that consume the MLP output, which halves the deviation of attention alone (\wPrevGaze{} to \wDownGaze{}).

\begin{table}[t]
\centering
\small
\setlength{\tabcolsep}{4pt}
\resizebox{\columnwidth}{!}{%
\begin{tabular}{lccc}
\toprule
Action-attention weight & alone & $\times$ exact error & $\times$ novelty \\
\midrule
No reader weight & -- & 12.60 & 12.15$^\dagger$ \\
Current block & -- & 11.19 & 13.07$^\dagger$ \\
Previous step, all blocks (LAQ~\cite{laq}) & 9.44$^\dagger$ & 5.16 & 6.32$^\dagger$ \\
Previous step, downstream blocks & 4.87$^\dagger$ & 3.33 & \textbf{3.91}$^\dagger$ \\
\midrule
Same step, downstream (reference run) & 4.63 & 3.22 & -- \\
\bottomrule
\end{tabular}}
\caption{Cosmos Policy 2B, 30\% of the 1764 tokens exact in every block and denoising step, 8 held-out observations; mean action deviation in dense-seed units (lower is better). Rows change the attention weight; columns use it alone or multiply it by a per-token score. $^\dagger$Uses only quantities available at inference. The last row reads the attention of a dense reference run and is an upper reference.}
\label{tab:wam}
\end{table}

Novelty is thus the per-token signal in both backbones. When the reader runs after the dispatched layers, as in $\pi_{0.5}$, novelty alone outperforms exact error; when the reader runs alongside, novelty sharpens the aligned attention at the cost of one narrow product, recovering most of what exact error would add (\wDownNov{} versus \wDownPerf{}, from \wDownGaze{} for attention alone).

\section{Implementing one decision per MLP}
\subsection{Drafts}
For each of the 17 live layers we fit rank-256 drafts that minimize the output error on calibration activations. For gate and up, $\tilde W = E D$ with $E = L Q_r$ and $D = Q_r^\top L^{-1} W$, where $L L^\top = W W^\top$ and $Q_r$ spans the top eigenvectors of $L^\top \Sigma_x L$; this is the output-optimal projection of~\cite{swiftsvd} computed in the input space. For down, $\tilde W = W V_r V_r^\top$ with $V_r$ the top eigenvectors of the output covariance. A rank-256 MLP costs $0.14$ of the original arithmetic.

\subsection{Which tokens to compute exactly}
At each layer, the $k = \mathrm{round}(f n)$ tokens with the highest score are computed exactly and the rest take the draft. A fixed $k$ keeps tensor shapes static. We compare scores that are available before the products:
\begin{itemize}
\item \textbf{Input residual}~\cite{gsi}: $\rho = \|y - y E F\| / \|y\|$, where $y$ is the MLP input and $F$ the reconstruction map of the gate draft.
\item \textbf{Output sketch}: $\|y (W_{\text{gate}} - \tilde W_{\text{gate}}) S\|$ with a $16384\times32$ Gaussian $S$. The narrow matrix $(W - \tilde W)S$ is precomputed, so the error of the draft is estimated before the draft is formed.
\item \textbf{Their union}: the larger of the two percentile ranks.
\end{itemize}
As references we evaluate a random choice, a static per-position score from calibration, and a \emph{perfect verifier} that computes both the exact and the low-rank MLP for every token and ranks tokens by the exact error of the MLP output. The perfect verifier is charged no cost.

\subsection{One decision per layer}
All scores above depend only on the MLP input $y$. We therefore decide once per token and layer and route the whole MLP. One product $h = yM$ with
\begin{equation}
M = \big[\,E_{\text{gate}}\;\big|\;E_{\text{up}}\;\big|\;(W_{\text{gate}} - \tilde W_{\text{gate}})S\,\big] \in \mathbb{R}^{2048\times544}
\end{equation}
returns both draft coefficients and the output sketch; the input residual adds one $256\times2048$ product (Algorithm~1).

\begin{figure}[t]
\small
\fbox{\begin{minipage}{0.95\columnwidth}
\textbf{Algorithm 1.} Dispatch at one layer.\\[2pt]
\emph{Offline:} $M$, $F$, $D_{\text{gate}}$, $D_{\text{up}}$, $E_{\text{down}}$, $D_{\text{down}}$.\\
\emph{Input:} MLP inputs $Y\in\mathbb{R}^{n\times2048}$, exact fraction $f$.
\begin{enumerate}\setlength{\itemsep}{0pt}
\item $H \leftarrow YM$ \hfill (one $2048\times544$ product)
\item $\rho_i \leftarrow \|y_i - h^{\text{gate}}_i F\| / \|y_i\|$ for every token $i$
\item $X \leftarrow$ the $k=\mathrm{round}(fn)$ tokens with the largest $\rho$; $L \leftarrow$ the rest
\item $Z_X \leftarrow \mathrm{MLP}(Y_X)$ \hfill (original weights)
\item $Z_L \leftarrow \big(\mathrm{GELU}(H^{\text{gate}}_L D_{\text{gate}}) \odot H^{\text{up}}_L D_{\text{up}}\big)\,E_{\text{down}} D_{\text{down}}$
\item scatter $Z_X$ and $Z_L$ back to token order
\end{enumerate}
\end{minipage}}
\end{figure}

Because $k$ is fixed, every tensor in the prefix has a static shape, so the whole dispatched prefix can be captured as one CUDA graph. A threshold gate, as in earlier token-level gating~\cite{ssi,gsi}, makes the split sizes data-dependent. Tokens that take the draft continue with $\mathrm{GELU}(h_{\text{gate}} D_{\text{gate}}) \odot (h_{\text{up}} D_{\text{up}})$ followed by the low-rank down projection, and only their 2048-wide rows are gathered and scattered. The remaining tokens run the original MLP. The relative MLP arithmetic is
\begin{equation}
\tfrac13\Big[\tfrac{2r + 32 + r}{16384} + (1-f)\big(\tfrac{2r}{2048} + \tfrac{r}{2048} + \tfrac{r}{16384}\big) + 3f\Big],
\end{equation}
which is \onceOps{20}, \onceOps{30}, \onceOps{40} and \onceOps{60} for $f = 0.2, 0.3, 0.4, 0.6$.

Deciding per product instead lets gate, up and down choose different tokens, but then the 16384-wide intermediate rows must be split and merged. On our GPU, copying these rows back costs \moveCopy{}\% of a full gate product, and gathering the down input and computing its norm cost \moveSel{}\% and \moveNorm{}\% of a full down product. The exact path is also inefficient at few rows: computing 20\% of the rows takes \rowsFrac{}\% of the full product's time.

\section{From MLP arithmetic to robot actions}

\subsection{Deciding once turns arithmetic into time}
\begin{figure*}[t]
\centering
\includegraphics[width=\textwidth]{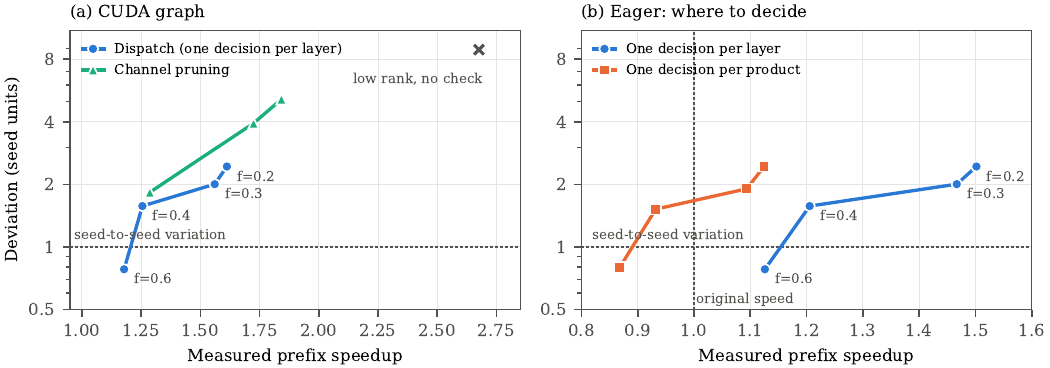}
\caption{Measured prefix speedup versus action deviation on held-out set B (deviation from Table~\ref{tab:main}; time on AC power). (a) Every arm captured as a CUDA graph: dispatch traces a front from \onceX{60}$\times$ at deviation \onceDev{60} to \onceX{20}$\times$ at \onceDev{20}; channel pruning and the unchecked low-rank prefix for reference. (b) Eager execution, same score and exact fraction: deciding once per token and layer instead of once per product moves the whole front to the right.}
\label{fig:tradeoff}
\end{figure*}

Figure~\ref{fig:tradeoff}b isolates where the decision is made. With the same score and the same exact fraction, one decision per product reaches \perX{20}, \perX{30}, \perX{40} and \perX{60}$\times$ in eager execution, below 1 for $f\ge0.4$, while one decision per layer reaches \eagerOnce{20}, \eagerOnce{30}, \eagerOnce{40} and \eagerOnce{60}$\times$. Episode-bootstrap intervals for the deviation ratio include 1 for $f\le0.4$. At $f{=}0.6$ the per-layer decision is 7\% closer. The kernel measurements identify a cost of per-product dispatch: per-product decisions must split and merge the 16384-wide intermediate of every layer.

Figure~\ref{fig:tradeoff}a compares the deployable form, with every arm captured as a CUDA graph. Dispatch reaches \onceX{20}, \onceX{30}, \onceX{40} and \onceX{60}$\times$ at $f = 0.2, 0.3, 0.4, 0.6$, and the unchecked low-rank prefix \noneX{}$\times$, the ceiling for rank-256 drafts on this device. Against channel pruning, dispatch at $f{=}0.4$ runs close to the speed of keeping 75\% of channels (\onceX{40} versus \chX{75}$\times$) at lower deviation (\onceDev{40} versus \chDev{75}), and between these two operating points the dispatch front lies below channel pruning: at $f{=}0.3$ dispatch runs at \onceX{30}$\times$ with deviation \onceDev{30}, while the line joining the 75\% and 50\% channel-retention measurements (\chX{50}$\times$, \chDev{50}) passes \chInterp{} at the same speed. The arithmetic advantage of dispatch is larger than its time advantage because the exact path runs on few rows: computing 20\% of the rows of a product takes \rowsFrac{}\% of its time. Pruning and dispatch act on different axes of the same MLP, so the exact path of dispatch can itself be a pruned MLP.

\subsection{Fidelity at a fixed arithmetic budget}

Table~\ref{tab:main} compares dispatch with three static compressions of the same MLP: a fixed rank of 1024, a rank allocated across the 51 products by output energy, and channel pruning ranked by activation energy and output weight norm on calibration data, a static form of the MLP channel pruning that AdaVLA and EcoVLA apply to $\pi_{0.5}$~\cite{adavla,ecovla}. At half the MLP arithmetic, dispatch deviates by \onceDev{40} seed units against \chDev{50} for channel pruning, and \staticGapLo{}--\staticGapHi{}$\times$ less than every static compression at equal or higher arithmetic. With 60\% of the tokens exact, the deviation falls to \onceDev{60}, below the policy's own seed-to-seed variation, while retaining 75\% of channels gives \chDev{75}. The test itself is what makes the draft usable: the same rank-256 drafts applied to all tokens deviate by \noneDev{}.

\begin{table}[t]
\centering
\small
\setlength{\tabcolsep}{4pt}
\resizebox{\columnwidth}{!}{%
\begin{tabular}{lcccc}
\toprule
 & MLP ops & Prefix & Deviation & Gripper \\
Method & (rel.) & speedup & (seed units) & agree.\ (\%) \\
\midrule
Original prefix & 1.00 & 1.00 & 0 & 100.00 \\
Low rank, no check & 0.14 & 2.68 & 8.88 & 93.02 \\
\midrule
Fixed rank 1024 & 0.56 & -- & 4.26 & 96.92 \\
Energy-allocated rank & 0.56 & -- & 4.05 & 97.38 \\
Channel pruning, keep 50\% & 0.50 & 1.72 & 3.93 & 97.18 \\
Channel pruning, keep 75\% & 0.75 & 1.29 & 1.83 & 98.75 \\
\midrule
\textbf{Dispatch}, $f{=}0.2$ & 0.32 & \textbf{1.61} & \textbf{2.44} & 98.37 \\
\textbf{Dispatch}, $f{=}0.3$ & 0.41 & \textbf{1.56} & \textbf{2.00} & 98.92 \\
\textbf{Dispatch}, $f{=}0.4$ & 0.49 & \textbf{1.26} & \textbf{1.57} & 99.15 \\
\textbf{Dispatch}, $f{=}0.6$ & 0.67 & \textbf{1.18} & \textbf{0.78} & 99.60 \\
\bottomrule
\end{tabular}}
\caption{Held-out set B (30 observations, unseen tasks and initial states, 4 noise seeds each). Deviation is the RMS pose difference to the original policy under identical noise, in units of the policy's own seed-to-seed variation. Speedups are whole-prefix GPU times with every arm captured as a CUDA graph and replayed in interleaved order; fixed and energy-allocated rank were evaluated for fidelity only.}
\label{tab:main}
\end{table}

\subsection{What the policy gains at the 30\% budget}
Table~\ref{tab:decision} times one whole control call. At $f=0.3$ the prefix falls from \decPrefDense{} to \decPrefThirty{}\,ms and the decision from \decDense{} to \decThirty{}\,ms (\decXThirty{}$\times$); at $f=0.6$ the decision takes \decSixty{}\,ms (\decXSixty{}$\times$). Image encoding and the action expert are unchanged, so they bound the whole-decision gain: with a free prefix the call would still take about \(\decImage{}+\decExpert{}\)\,ms.

\begin{table}[t]
\centering
\small
\setlength{\tabcolsep}{3pt}
\resizebox{\columnwidth}{!}{%
\begin{tabular}{lcccc}
\toprule
 & Image & Prefix & Action & Whole decision \\
 & encoding & & expert (10 steps) & p50 / p95 \\
\midrule
Original policy & 34.8 & 94.5 & 62.0 & 194.1 / 198.6 \\
NoveltyGate, $f{=}0.3$ & 34.8 & 60.1 & 62.0 & 159.6 / 163.8 \\
NoveltyGate, $f{=}0.6$ & 34.8 & 80.7 & 62.2 & 180.7 / 184.6 \\
\bottomrule
\end{tabular}}
\caption{One batch-one control call on the RTX 5060 Laptop GPU (115\,W), milliseconds. Image encoding, prefix and the ten action-expert steps are each captured as CUDA graphs; the image and expert graphs are shared by all arms, and each process holds the original prefix and one dispatch prefix, three observations $\times$ ten interleaved repetitions. Whole decision includes host-side input processing and action readback.}
\label{tab:decision}
\end{table}

On ten paired closed-loop episodes (five held-out LIBERO tasks, two new initial states each), the original policy and dispatch at $f=0.3$ each succeed in \clThirty{}. Dispatch at $f=0.6$ succeeds in \clSixty{}, and sending every token through the same draft succeeds in \clNone{}. These outcomes establish that selective exact computation can recover successful rollouts in this sample. A larger study is needed to estimate changes in success rate.

\section{Related Work}
\paragraph{Token-level low-rank gating.} Skyline Subspace Inference (SSI)~\cite{ssi} tracks an activation subspace online and computes a token through the full MLP projections only when the relative residual of its input outside that subspace exceeds a threshold; it reports a 2.01$\times$ speedup for the MLP projections of language models on an MI300X. Gated Subspace Inference (GSI)~\cite{gsi} extends the same gate from the MLP to all linear maps of a layer, caches $W V_k$, bounds the output error with a spectral norm, and reports speedups computed from gate statistics. Both target language-model inference. We adopt their residual as a score and contribute what the VLA prefix adds to the picture: evidence that this input-side test selects better tokens than even a perfect verifier of output error, judged by the actions a policy takes; a single decision per token and layer for a gated MLP, with output-optimal drafts and a fixed quota that keeps every shape static; the measurement that per-product decisions lose the speedup on a GPU; and closed-loop evaluation.

\paragraph{Learned conditional computation.} Fast Forward~\cite{fastforward} trains a predictor and an error compensator to sparsify prefill FFNs per block. CoLT5~\cite{colt5} and Mixture-of-Depths~\cite{mod} route a top-$k$ set of tokens through a heavy branch with trained routers, and GateSkip~\cite{gateskip} learns residual gates under a token budget. Dispatch is training-free and keeps the original weights for the exact tokens.

\paragraph{Reconstruction residuals as importance.} ApET~\cite{apet} compresses the visual tokens of VLMs by reconstructing every token as a linear combination of a few base tokens and keeping the tokens that the base explains worst. This residual of the input representation is close in spirit to our novelty score, which measures the part of a token's input outside the calibrated subspace. The quantity our perfect verifier ranks is different: the error of approximating a layer's output, which we compare directly with calibrated input novelty.

Deputy~\cite{deputy} uses a learned module to route LLM tokens through full or low-rank computation. Our fixed-quota comparison isolates the effect of the score on robot actions and measures the cost of placing the gate inside versus before the MLP.

\paragraph{Action attention as importance.} LAQ~\cite{laq} quantizes world--action models to 4 bits and protects tokens with a gaze computed from action attention, reusing attention from the previous control or denoising step and summing it over layers. NoveltyGate addresses a different problem, routing tokens between a draft and the exact MLP, and contributes a different reader: attention aligned with the blocks that consume the MLP output, combined with input novelty. In the same router, this rule beats LAQ's all-layer gaze on every test observation, including when LAQ's gaze is paired with exact error (Sec.~\ref{sec:wam}).

\paragraph{Cheap preview, exact refinement.} SVD-softmax~\cite{svdsoftmax} previews output logits with a truncated SVD and recomputes the top rows at full width; dual-module inference~\cite{dualmodule} predicts insensitive neurons with a small module; Davis and Arel~\cite{davis} use a low-rank product to predict which ReLU units are zero. These methods select outputs within a product; dispatch selects tokens across a whole MLP, which keeps the exact path a dense product.

\paragraph{Checking approximate computation.} Rumba~\cite{rumba} and MITHRA~\cite{mithra} decide per invocation whether an approximate accelerator's output is acceptable. ApproxEigen~\cite{approxeigen} checks approximate matrix-vector products with column-sum projections, $c^\top(Ax) = (c^\top A)x$, after computing them, and rolls back. Freivalds' test~\cite{freivalds}, algorithm-based fault tolerance~\cite{abft}, Slalom~\cite{slalom} and FP-Sketch~\cite{fpsketch} verify products by random projection for correctness, security or fault recovery, and randomized residual estimators certify reduced models~\cite{smetana}. We borrow the projection as a tool and find that, for a policy, the product's own error is not the quantity to check.

\paragraph{Accelerating VLAs.} AdaVLA~\cite{adavla} and EcoVLA~\cite{ecovla} prune MLP channels of $\pi_{0.5}$ adaptively; IAprune~\cite{iaprune} removes visual tokens that do not align with the manipulation; HoloQ-VLA~\cite{holoq} quantizes the backbone and action head of VLAs to W4A4. Other work shortens the denoising chain~\cite{probeflow}, speculates across calls~\cite{flash,srvla}, or optimizes the kernels of the whole pipeline~\cite{realtimevla}. Dispatch keeps every token and every channel and changes how precisely each token is computed, so it is orthogonal to channel pruning, token pruning and quantization: the exact path can itself be pruned or quantized, and the dispatched tokens can be a pruned set.

\section{Conclusion}
Input novelty decides which tokens a robot policy should compute exactly. In the $\pi_{0.5}$ VLA prefix it outperforms exact knowledge of every token's local MLP error, because it tracks where the action expert reads. In the Cosmos Policy WAM, where that reader runs alongside the dispatched layers, novelty multiplied by the previous step's downstream action attention cuts deviation to a third, and novelty improves on the attention on every test observation. A single fixed-quota decision before the wide MLP products turns the rule into measured end-to-end time with stock PyTorch kernels.

\appendix
\section{Exploratory reader proxy}
A follow-up study asks whether a reader signal available during prefill can improve an inexpensive error estimate. We average post-RoPE action queries separately for each head and layer on 15 calibration observations. At inference, the prototype scores the \emph{current} layer's prefix keys by scaled dot product, followed by a prefix-only softmax and a mean over heads. The score is a proxy for downstream dependence. The current MLP can change only later cache entries. Related Q-Filters use calibrated query directions for LLM cache selection~\cite{qfilters}.

We test five held-out tasks with two fresh initial states each, using only the first three decisions per rollout (30 observations, four paired noise seeds, $f=0.3$). Mean deviations are 1.64 for novelty, 2.25 for the 32-dimensional gate-projection error sketch, 1.57 for the reader proxy, and 1.31 for their product. Replacing the two camera images used to derive the reader weights with images from the other initial state at the same task and decision time gives 1.62; the evaluated policy still receives its own observation. The product improves on novelty in 20 of 30 observations, so a reader estimate formed during prefill already adds to novelty in $\pi_{0.5}$, the same pairing that works best in Cosmos (Sec.~\ref{sec:wam}).

\small
\bibliographystyle{unsrtnat}
\bibliography{refs}
\end{document}